\documentclass{article}

\usepackage[final]{nips_2017}

\usepackage[utf8]{inputenc} 
\usepackage[T1]{fontenc}    
\usepackage{hyperref}       
\usepackage{url}            
\usepackage{booktabs}       
\usepackage{amsfonts}       
\usepackage{nicefrac}       
\usepackage{microtype}      

\usepackage{amssymb}
\usepackage{amsmath}
\usepackage{graphicx}
\usepackage{caption}
\usepackage{subcaption}
\usepackage{algorithm}
\usepackage{algpseudocode}
\usepackage{amsthm}
\usepackage{wrapfig}
\usepackage{float}

\algtext*{EndWhile}
\algtext*{EndIf}
\algtext*{EndFor}
\algtext*{Until}

\newcommand{\E}{\mathbb{E}}

\title{Leave no Trace: Learning to Reset for Safe and Autonomous Reinforcement Learning}

\author{
Benjamin Eysenbach\thanks{Work done as a member of the Google Brain Residency Program (\texttt{\href{https://research.google.com/teams/brain/residency/}{g.co/brainresidency}})}\textsuperscript{\;\; $\dagger$}, Shixiang Gu\textsuperscript{$\dagger$ $\ddagger$ $\dagger\dagger$}, Julian Ibarz\textsuperscript{$\dagger$}, Sergey Levine\textsuperscript{$\dagger$ $\ddagger\ddagger$} \\
\textsuperscript{$\dagger$}Google Brain \\
\textsuperscript{$\ddagger$}University of Cambridge \\
\textsuperscript{$\dagger\dagger$}Max Planck Institute for Intelligent Systems \\
\textsuperscript{$\ddagger\ddagger$}UC Berkeley \\
\texttt{\{eysenbach,shanegu,julianibarz,slevine\}@google.com} \\
\vspace{-3em}
}

\begin{document}

\maketitle

\begin{abstract}
  Deep reinforcement learning algorithms can learn complex behavioral skills, but real-world application of these methods requires a large amount of experience to be collected by the agent. In practical settings, such as robotics, this involves repeatedly attempting a task, resetting the environment between each attempt. However, not all tasks are easily or automatically reversible. In practice, this learning process requires extensive human intervention. In this work, we propose an autonomous method for safe and efficient reinforcement learning that simultaneously learns a forward and reset policy, with the reset policy resetting the environment for a subsequent attempt. By learning a value function for the reset policy, we can automatically determine when the forward policy is about to enter a non-reversible state, providing for uncertainty-aware safety aborts. Our experiments illustrate that proper use of the reset policy can greatly reduce the number of manual resets required to learn a task, can reduce the number of unsafe actions that lead to non-reversible states, and can automatically induce a curriculum.\footnote{Videos of our experiments: \url{https://sites.google.com/site/mlleavenotrace/}}
  \vspace{-1em}
\end{abstract}

\vspace{-0.2em}
\section{Introduction}
Deep reinforcement learning (RL) algorithms have the potential to automate acquisition of complex behaviors in a variety of real-world settings. Recent results have shown success on games~(\cite{mnih2013playing}), locomotion~(\cite{schulman2015high}), and a variety of robotic manipulation skills~(\cite{pinto2017learning, schulman2016learning, gu2017deep}). However, the complexity of tasks achieved with deep RL in simulation still exceeds the complexity of the tasks learned in the real world. Why have real-world results lagged behind the simulated accomplishments of deep RL algorithms?

One challenge with real-world application of deep RL is the scaffolding required for learning: a bad policy can easily put the system into an unrecoverable state from which no further learning is possible. For example, an autonomous car might collide at high speed, and a robot learning to clean glasses might break them. Even in cases where failures are not catastrophic, some degree of human intervention is often required to reset the environment between attempts (e.g.~\cite{Chebotar2017PathIG}).

Most RL algorithms require sampling from the initial state distribution at the start of each episode. On real-world tasks, this operation often corresponds to a human resetting the environment after every episode, an expensive solution for complex environments. Even when tasks are designed so that these resets are easy (e.g.~\cite{levine2016learning} and~\cite{gu2017deep}), manual resets are necessary when the robot or environment breaks (e.g.~\cite{gandhi2017learning}). The bottleneck for learning many real-world tasks is not that the agent collects data too slowly, but rather that data collection stops entirely when the agent is waiting for a manual reset.
To avoid manual resets caused by the environment breaking, humans often add negative rewards to dangerous states and intervene to prevent agents from taking dangerous actions. While this works well for simple tasks, scaling to more complex environments requires writing large numbers of rules for types of actions the robot should avoid. For example, a robot should avoid hitting itself, except when clapping. One interpretation of our method is as automatically learning these safety rules.
Decreasing the number of manual resets required to learn to a task is important for scaling up RL experiments outside simulation, allowing researchers to run longer experiments on more agents for more hours.

We propose to address these challenges by forcing our agent to ``leave no trace.'' The goal is to learn not only how to do the task at hand, but also how to undo it. The intuition is that the sequences of actions that are reversible are safe; it is always possible to undo them to get back to the original state. This property is also desirable for continual learning of agents, as it removes the requirements for manual resets. In this work, we learn two policies that alternate between attempting the task and resetting the environment. By learning how to reset the environment at the end of each episode, the agent we learn requires significantly fewer manual resets. Critically, our value-based reset policy restricts the agent to only visit states from which it can return, intervening to prevent the forward policy from taking potentially irreversible actions.
The set of states from which the agent knows how to return grows over time, allowing the agent to explore more parts of the environment as soon as it is safe to do so.

The main contribution of our work is a framework for continually and jointly learning a reset policy in concert with a forward task policy. We show that this reset policy not only automates resetting the environment between episodes, but also helps ensure safety by reducing how frequently the forward policy enters unrecoverable states. Incorporating uncertainty into the value functions of both the forward and reset policy further allows us to make this process risk-aware, balancing exploration against safety. Our experiments illustrate that this approach reduces the number of ``hard'' manual resets required during learning of a variety of simulated robotic skills.

\vspace{-0.2em}
\section{Related Work}
\label{sec:related-work}

Our method builds off previous work in areas of safe exploration, multiple policies, and automatic curriculum generation. Previous work has examined safe exploration in small MDPs. \cite{moldovan2012risk} examine risk-sensitive objectives for MDPs, and propose a new objective of which minmax and expectation optimization are both special cases. \cite{moldovan2012safe} consider safety using ergodicity, where an action is safe if it is still possible to reach every other state after having taken that action.
These methods are limited to small, discrete MDPs where exact planning is straightforward.
Our work includes a similar notion of safety, but can be applied to solve complex, high-dimensional tasks.

Previous work has also used multiple policies for safety and for learning complex tasks. 
\cite{han2015learning} learn a sequence of forward and reset policies to complete a complex manipulation task. Similar to \cite{han2015learning}, our work learns a reset policy to undo the actions of the forward policy. While \cite{han2015learning} engage the reset policy when the forward policy fails, we preemptively predict whether the forward policy will fail, and engage the reset policy before allowing the forward policy to fail.
Similar to our approach, \cite{richter2017safe} also propose to use a safety policy that can trigger an ``abort'' to prevent a dangerous situation. However, in contrast to our approach, \cite{richter2017safe} use a heuristic, hand-engineered reset policy, while our reset policy is learned simultaneously with the forward policy. \cite{kahn17} uses uncertainty estimation via bootstrap to provide for safety. Our approach also uses bootstrap for uncertainty estimation, but unlike our method, \cite{kahn17} does not learn a reset or safety policy.

Learning a reset policy is related to curriculum generation: the reset controller is engaged in increasingly distant states, naturally providing a curriculum for the reset policy.
Prior methods have studied curriculum generation by maintaining a separate goal setting policy or network \citep{sukhbaatar2017intrinsic,matiisen2017teacher,held2017automatic}. In contrast to these methods, we do not set explicit goals, but only allow the reset policy to abort an episode.
When learning the forward and reset policies jointly, the training dynamics of our reset policy resemble those of reverse curriculum generation \citep{florensa2017reverse}, but in reverse.
In particular, reverse curriculum learning can be viewed as a special case of our method: our reset policy is analogous to the learner in the reverse curriculum, while the forward policy plays a role similar to the initial state selector. However, reverse curriculum generation requires that the agent can be reset to any state (e.g., in a simulator), while our method is specifically aimed at streamlining real-world learning, through the use of uncertainty estimation and early aborts.

\vspace{-0.2em}
\section{Preliminaries}

In this section, we discuss the episodic RL problem setup, which motivates our proposed joint learning of forward and reset policies. RL considers decision making problems that consist of a state space $\mathcal{S}$, action space $\mathcal{A}$, transition dynamics $P(s' \mid s, a)$, an initial state distribution $p_0(s)$, and a scalar reward function $r(s,a)$. In episodic, finite horizon tasks, the objective is to find the optimal policy $\pi^*(a \mid s)$ that maximizes the expected sum of $\gamma$-discounted returns, $\E_\pi[ \sum_{t=0}^T \gamma^t r(s_t, a_t)]$, where $s_0 \sim p_0$, $a_t \sim \pi(a_t | s_t)$, and $s_{t+1} \sim P(s'\mid s, a)$.

Typically, the RL training routines involve iteratively sampling new episodes, where at the end of each episode, a new starting state $s_0$ is sampled from a given initial state distribution $p_0$. In practical applications, such as robotics, this procedure effectively involves executing some hard-coded reset policy or human interventions to manually reset the agent. Our work is aimed at avoiding these manual resets by learning an additional reset policy that satisfies the following property: when the reset policy is executed from any state, the distribution over final states matches the initial state distribution $p_0$. If we learn such a reset policy, then the agent never requires querying the black-box distribution $p_0$ and can continually learning on its own.

\vspace{-0.2em}
\section{Continual Learning with Joint Forward-Reset Policies}
\label{sec:method}

Our method for continual learning relies on jointly learning a \emph{forward policy} and \emph{reset policy}, and using \emph{early aborts} to avoid manual resets.
The forward policy aims to maximize the task reward, while the reset policy takes actions to reset the environment. Both have the same state and action spaces, but are given different reward objectives. The forward policy reward $r_f(s,a)$ is the usual task reward given by the environment.
The reset policy reward $r_r(s)$ is designed to be large for states with high density under the initial state distribution. For example, in locomotion experiments, the reset reward is large when the agent is standing upright. To make this set-up applicable for solving the task, we make the weak assumption on the task environment that there exists a policy that can reset from at least one of the reachable states with maximum reward in the environment. This assumption ensures that it is possible to solve the task without any manual resets.

We choose off-policy actor-critic methods as the base RL algorithm~\citep{silver2014deterministic,lillicrap2015continuous}, since its off-policy learning allows sharing of the experience between the forward and reset policies. Additionally, the Q-functions can be used to signal early aborts. Our methods can also be used directly with any other Q-learning methods~(\citep{watkins1992q,mnih2013playing,gu2017deep,amos2016input,metz2017discrete}).

\subsection{Early Aborts}
\label{sec:early-aborts}
The reset policy learns how to transition from the forward policy's final state back to an initial state. However, in challenging domains with irreversible states, the reset policy may be unable to reset from some states, and a costly ``hard'' reset may be required. The process of learning the reset policy offers us a natural mechanism for reducing these hard resets. We observe that, for states that are irrecoverable, the value function of the reset policy will be low. We can therefore use this value function (or, specifically, its Q-function) as a metric to determine when to terminate the forward policy, performing an \emph{early abort}.

Before an action proposed by the forward policy is executed in the environment, it must be ``approved'' by the reset policy. In particular, if the reset policy's Q value for the proposed action is too small, then an early abort is performed: the proposed action is not taken and the reset policy takes control. 
Formally, early aborts restrict exploration to a `safe' subspace of the MDP.
Let $\mathcal{E} \subseteq \mathcal{S} \times \mathcal{A}$ be the set of (possible stochastic) transitions, and let $Q_{reset}(s, a)$ be the Q values of our reset policy at state $s$ taking action $a$. The subset of transitions $\mathcal{E}^* \in \mathcal{E}$ allowed by our algorithm is
\begin{equation}
\mathcal{E}^* \triangleq \{(s, a) \in \mathcal{E} \; | \; Q_{reset}(s, a) > Q_{min} \}
\end{equation}
Noting that $Q(s) \triangleq \max_{a \in \mathcal{A}} Q(s, a)$, we see that the states allowed under our algorithm $\mathcal{S}^* \subseteq \mathcal{S}$ are those states with at least one transition in $\mathcal{E}^*$:
\begin{equation}
\mathcal{S}^* \triangleq \{s \; | \; (s, a) \in \mathcal{E}^* \text{ for at least one } a \in \mathcal{A} \}
\end{equation}
For intuition, consider tasks where the reset reward is 1 if the agent has successfully reset, and 0 otherwise. In these cases, the reset Q function is the probability that the reset will succeed. Early aborts occur when this probability for the proposed action is too small.

Early aborts can be interpreted as a learned, dynamic, safety constraint, and a viable alternative for the manual constraints that are typically used for real-world robotic learning experiments. Early aborts promote safety by preventing the agent from taking actions from which it cannot recover. These aborts are dynamic because the states at which they occur change throughout training, as more and more states are considered safe. This can make it easier to learn the forward policy, by preventing it from entering unsafe states. We analyze this experimentally in Section~\ref{sec:early-aborts-experiment}.

\subsection{Hard Resets}
\label{sec:manual-resets}

Early aborts decrease the requirement for ``hard'' resets, but do not eliminate them, since an imperfect reset policy might still miss a dangerous state early in the training process.
However, it is challenging to identify whether it is possible for \emph{any} policy to reset from the current state.
Our approach is to approximate irreversible state identification with a necessary (but not sufficient) condition: we say we have reached an irreversible state if the reset policy fails to reset after $N$ attempts, where $N$ is a hyperparameter. Formally, we define a set of safe states $\mathcal{S}_{reset} \subseteq \mathcal{S}$, and say that we are in an irreversible state if the set of states visited by the reset policy over the past $N$ episodes is disjoint from  $\mathcal{S}_{reset}$. Increasing $N$ decreases the number of hard resets. However, when we are in an irreversible state, increasing $N$ means that we remain in that state (learning nothing) for more episodes. Section~\ref{sec:reset-attempts-experiment} empirically examines this trade-off. In practice, the setting of this parameter should depend on the cost of hard resets.

\subsection{Algorithm Summary}

\setlength{\textfloatsep}{1\baselineskip}
\begin{algorithm}[t!]
\caption{Joint Training}\label{alg:joint-training}
\begin{algorithmic}[1]
\Repeat
    \For{\texttt{max\_steps\_per\_episode}}
        \State $a \gets$ \Call{forward\_agent.act}{$s$}
        \If{\Call{reset\_agent.q}{$s, a$} $< Q_{min}$}
            \State{Switch to reset policy.} \Comment{Early Abort}
        \EndIf
        \State $(s, r) \gets$ \Call{environment.step}{$a$}
        \State Update the forward policy.
    \EndFor
    \For{\texttt{max\_steps\_per\_episode}}
        \State $a\gets$ \Call{reset\_agent.act}{$s$}
        \State $(s, r) \gets$ \Call{environment.step}{$a$}
        \State Update the reset policy.
    \EndFor
    \State Let $\mathcal{S}_{reset}^{i}$ be the final states from the last $N$ reset episodes.
    \If{$\mathcal{S}_{reset}^{i} \cap \mathcal{S}_{reset} = \emptyset$}
        \State $s \gets$ \Call{environment.reset}{\null} \Comment{Hard Reset}
    \EndIf
\Until{}
\end{algorithmic}
\end{algorithm}

Our full algorithm (Algorithm~\ref{alg:joint-training}) consists of alternately running a forward policy and reset policy. When running the forward policy, we perform an \emph{early abort} if the Q value for the reset policy is less than $Q_{min}$. Only if the reset policy fails to reset after $N$ episodes do we do a manual reset. 

\subsection{Value Function Ensembles}
\label{sec:ensembles}

The accuracy of the Q-value estimates of our policies affect learning and reset performance through early aborts. Q-values of the reset policy may not be good estimates of the true value function for previously-unseen states. To address this, we train Q-functions for both the forward and reset policies that provide uncertainty estimates. Several prior works have explored how uncertainty estimates can be obtained in such settings~\citep{gal2016dropout,osband2016deep}. We use the bootstrap ensemble in our method~\citep{osband2016deep}, though other techniques could be employed. In this approach, we train an ensemble of Q-functions, each with a different random initialization, which provides a distribution over Q-values at each state.

Given this distribution over Q values, we can propose three strategies for early aborts:
\begin{itemize}
    \item[] \emph{Optimistic Aborts:} Perform an early abort only if all the Q values are less than $Q_{min}$. Equivalently, do an early abort if $\max_{\theta} Q_{reset}^{\theta} (s, a) < Q_{min}$.
    \item[] \emph{Realist Aborts:} Perform an early abort if the mean Q value is less than $Q_{min}$.
    \item[] \emph{Pessimistic Aborts:} Perform an early abort if any of the Q values are less than $Q_{min}$. Equivalently, do an early abort if $\min_{\theta} Q_{reset}^{\theta} (s, a) < Q_{min}$.
\end{itemize}
We expect that optimistic aborts will provide better exploration at the cost of more hard resets, while pessimistic aborts should decrease hard resets, but may be unable to effectively explore. We empirically test this hypothesis in Appendix~\ref{sec:combining-ensemble-experiments}.

\vspace{-0.2em}
\section{Small-Scale Didactic Example}
\label{sec:gridworld-experiments}

\begin{wrapfigure}[28]{r}{0.4\textwidth}
    \captionsetup{width=0.35\textwidth}
    \centering
        \includegraphics[width=0.35\textwidth]{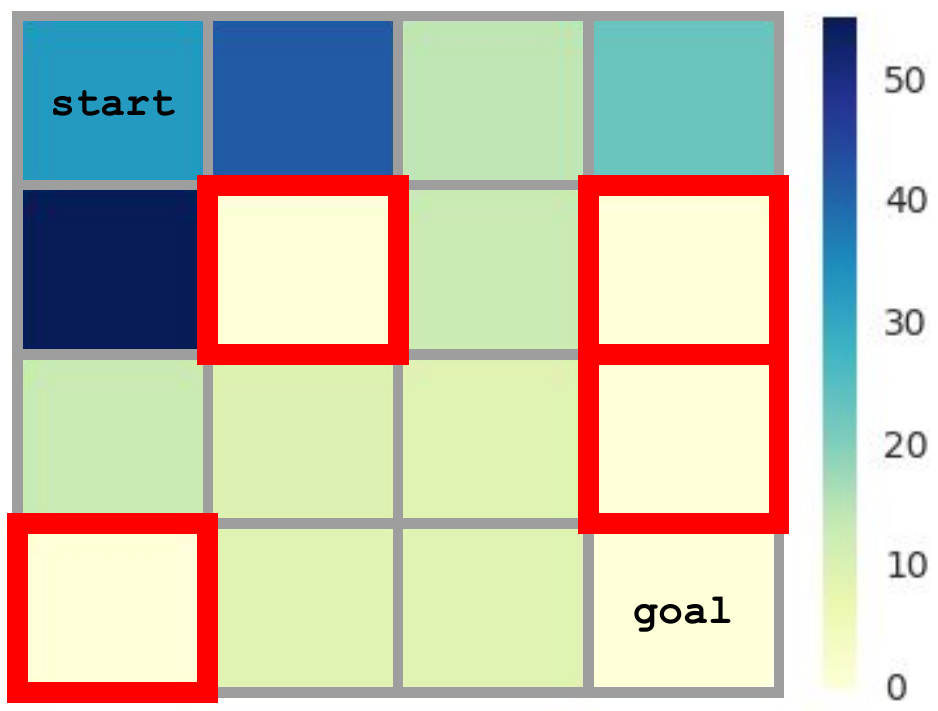}
        \caption{Early aborts in gridworld. \label{fig:where-are-early-aborts}}
        \vspace{1em}
        \includegraphics[width=0.35\textwidth]{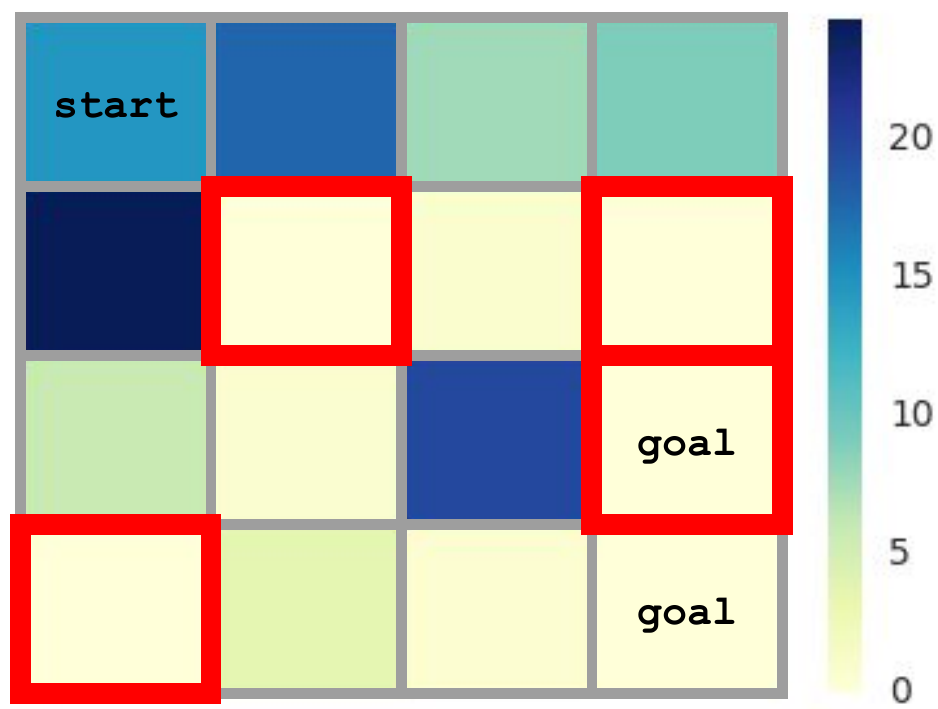}
        \caption{Early aborts with an \mbox{absorbing} goal. \label{fig:where-are-early-aborts-irreversible}}
\end{wrapfigure}

We first present a small didactic example to illustrate how our forward and reset policies interact and how cautious exploration reduces the number of hard resets. We first discuss the gridworld in Figure~\ref{fig:where-are-early-aborts}. The states with red borders are absorbing, meaning that the agent cannot leave them and must use a hard reset. The agent receives a reward of 1 for reaching the goal state, and 0 otherwise. The states are colored based on the number of early aborts triggered in each state. Note that most aborts occur next to the initial state, when the forward policy attempts to enter the absorbing state South-East of the start state, but is blocked by the reset policy.
In Figure~\ref{fig:where-are-early-aborts-irreversible}, we present a harder environment, where the task can be successfully completed by reaching one of the two goals, exactly one of which is reversible. The forward policy has no preference for which goal is better, but the reset policy successfully prevents the forward policy from entering the absorbing goal state, as indicated by the much larger early abort count in the blue-colored state next to the absorbing goal.

\begin{figure}[b!]
    \centering
    \begin{subfigure}[t]{0.5\textwidth}
        \centering
        \includegraphics[width=\textwidth]{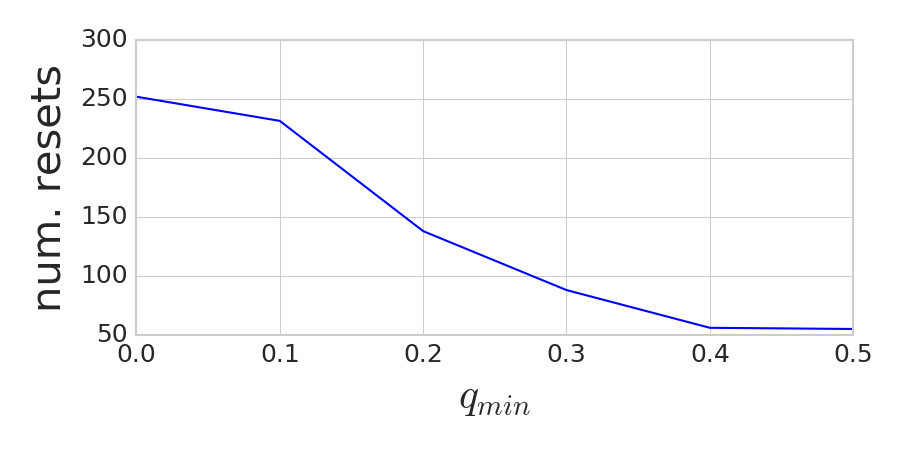}
    \end{subfigure}%
     ~ 
    \begin{subfigure}[t]{0.5\textwidth}
        \centering
        \includegraphics[width=\textwidth]{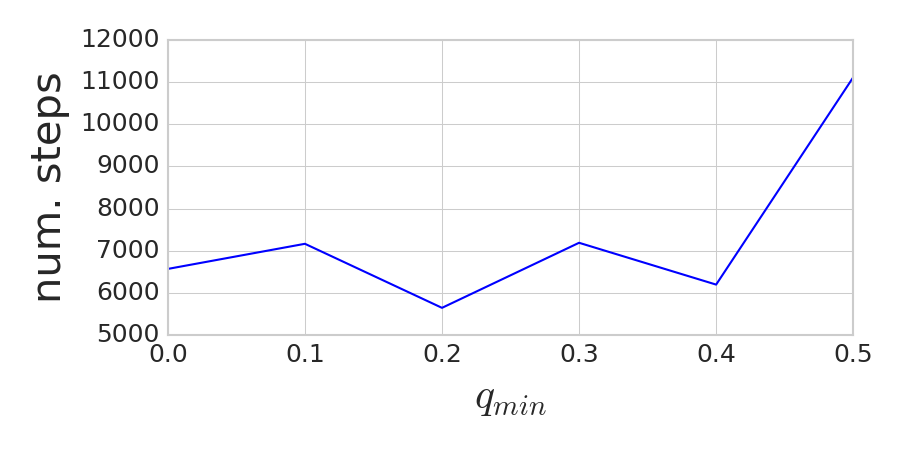}
    \end{subfigure}%
    \vspace{-0.5em}
    \caption{Early abort threshold: In our didactic example, increasing the early abort threshold causes more cautious exploration (left) without severely increasing the number of steps to solve (right). \label{fig:gridworld-qmin}}
    \vspace{-1em}
\end{figure}

Figure~\ref{fig:gridworld-qmin} shows how changing the early abort threshold to explore more cautiously reduces the number of failures. Increasing $Q_{min}$ from 0 to 0.4 reduced the number of hard resets by 78\% without increasing the number of steps to solve the task. In a real-world setting, this might produce a substantial gain in efficiency, as time spend waiting for a hard reset could be better spent collecting more experience. Thus, for some real-world experiments, increasing $Q_{min}$ can decrease training time even if it requires more steps to learn.

\vspace{-0.2em}
\section{Continuous Environment Experiments}
\label{sec:cts-experiments}

\begin{figure}[th!]
    \centering
    \begin{subfigure}[t]{0.19\textwidth}
        \centering
        \includegraphics[width=\textwidth]{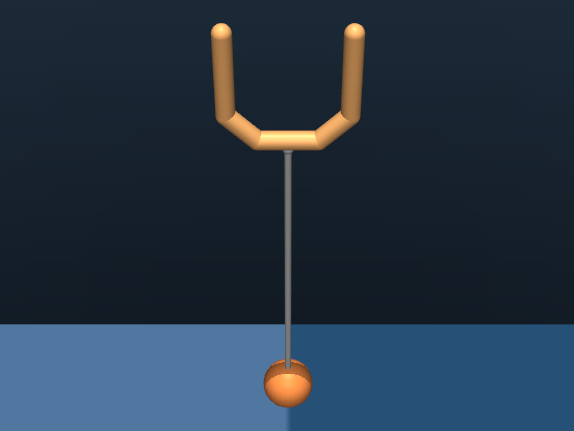}
        \caption*{Ball in Cup}
   \end{subfigure}
   ~ 
   \begin{subfigure}[t]{0.19\textwidth}
        \centering
        \includegraphics[width=\textwidth]{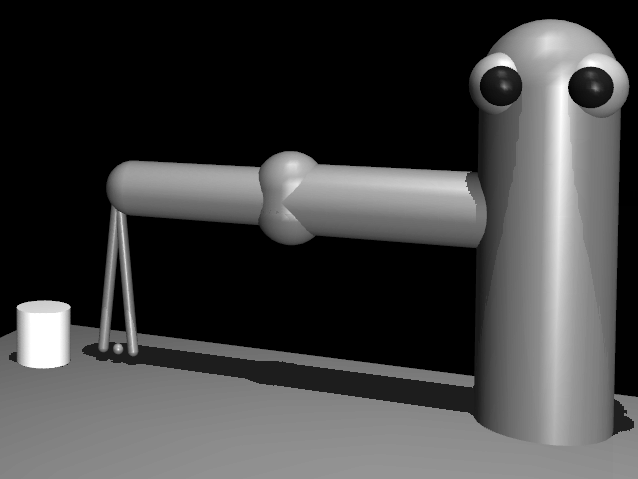}
        \caption*{Pusher}
   \end{subfigure}
   ~ 
   \begin{subfigure}[t]{0.19\textwidth}
        \centering
        \includegraphics[width=\textwidth]{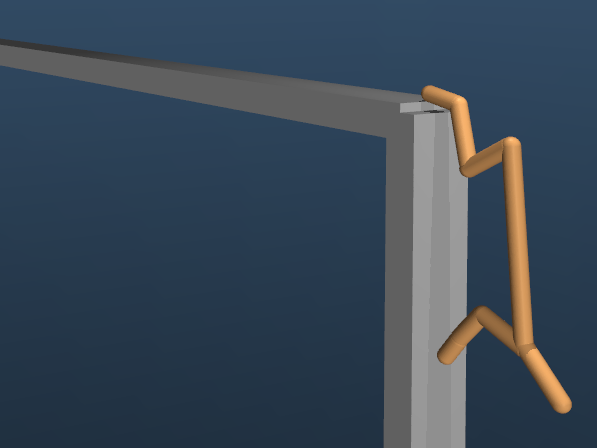}
        \caption*{Cliff Cheetah}
   \end{subfigure}
    ~ 
   \begin{subfigure}[t]{0.19\textwidth}
        \centering
        \includegraphics[width=\textwidth]{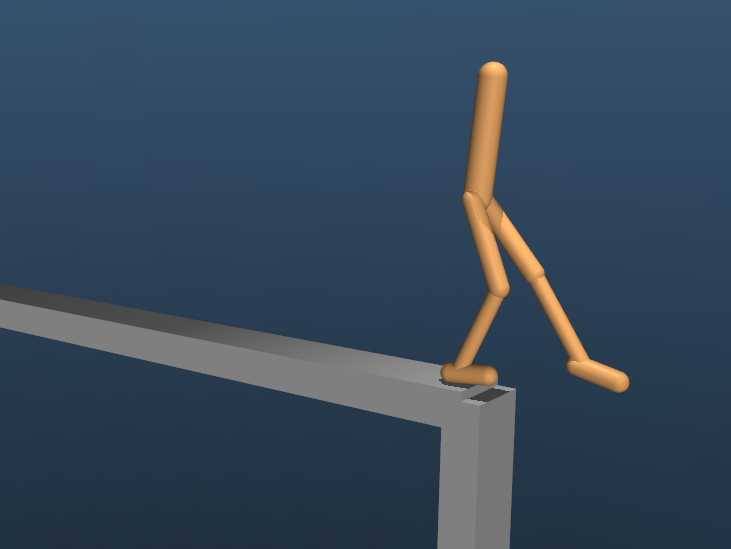}
        \caption*{Cliff Walker}
    \end{subfigure}
    ~ 
   \begin{subfigure}[t]{0.19\textwidth}
        \centering
        \includegraphics[width=\textwidth]{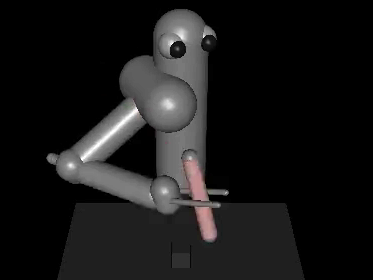}
        \caption*{Peg Insertion}
  \end{subfigure}
\end{figure}

In this section, we use the five complex, continuous control environments shown above to answer questions about our approach. While ball in cup and peg insertion are completely reversible, the other environments are not: the pusher can knock the puck outside its workspace and the cheetah and walker can jump off a cliff. Crucially, reaching these irreversible states does not terminate the episode, so the agent remains in the irreversible state until it calls for a hard reset. Additional plots and experimental details are in the Appendix.

\subsection{Why Learn a Reset Controller?}
\label{sec:forward-only-fail}

\begin{wrapfigure}[22]{r}{0.45\textwidth}
    \captionsetup{width=0.45\textwidth}
    \vspace{-1em}
    \centering
        \includegraphics[width=0.45\textwidth]{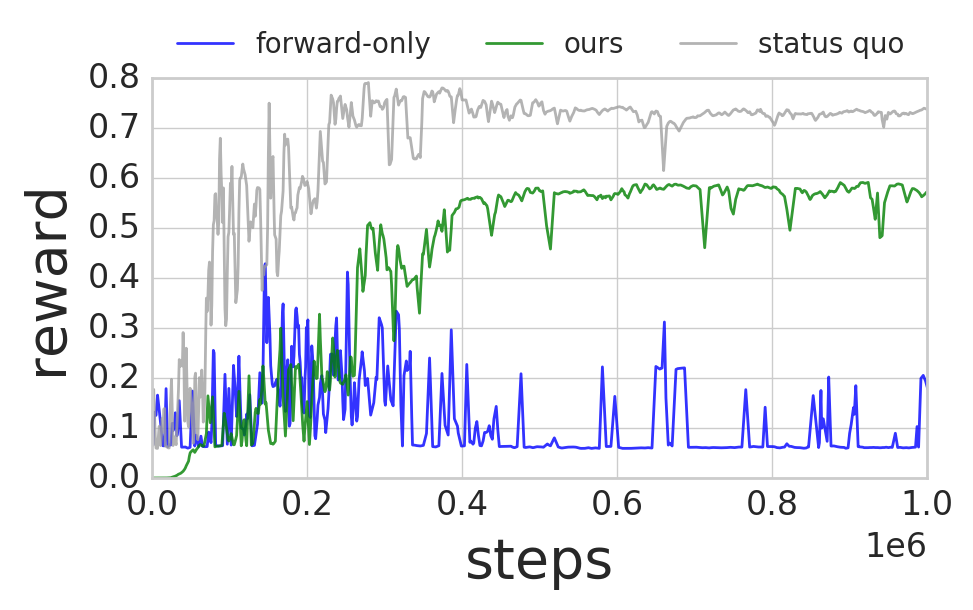}
        \caption{We compare our method to a non-episodic (``forward-only'') approach on ball in cup. Although neither uses hard resets, only our method learns to catch the ball. As an upper bound, we also show the ``status quo'' approach that performs a hard reset after episode, which is often impractical outside simulation. \label{fig:forward-only-fail}}
\end{wrapfigure}

One proposal for learning without resets is to run the forward policy until the task is learned. This ``forward-only'' approach corresponds to standard, fully online, non-episodic lifelong RL setting, commonly studied in the context of temporal difference learning (\cite{sutton1998reinforcement}).
We show that this approach fails, even on reversible environments where safety is not a concern.
We benchmarked the forward-only approach and our method on ball in cup, using no hard resets for either. Figure~\ref{fig:forward-only-fail} shows that our approach solves the task while the ``forward-only'' approach fails to learn how to catch the ball when initialized below the cup.
Once the forward-only approach catches the ball, it gets maximum reward by keeping the ball in the cup. In contrast, our method learns to solve this task by automatically resetting the environment after each episode, so the forward policy can practice catching the ball when initialized below the cup.
As an upper bound, we show policy reward for the ``status quo'' approach, which performs a hard reset after every episode. Note that the dependence on hard resets makes this third method impractical outside simulation.

\subsection{Does Our Method Reduce Manual Resets?}
\label{sec:learning-with-resets}

\begin{figure}[th!]
    \centering
    \vspace{-0.5em}
    \begin{subfigure}[t]{0.5\textwidth}
        \centering
        \includegraphics[width=\textwidth]{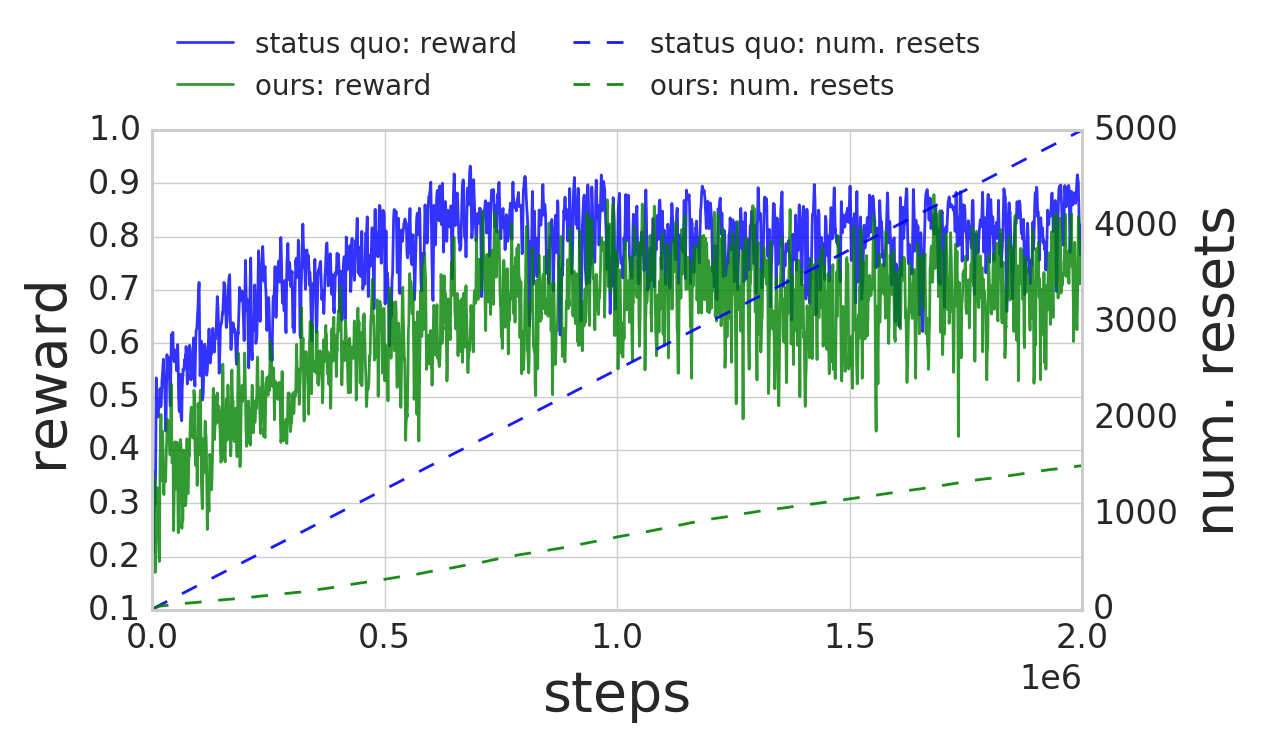}
        \caption*{Pusher}
    \end{subfigure}%
     ~ 
    \begin{subfigure}[t]{0.5\textwidth}
        \centering
        \includegraphics[width=\textwidth]{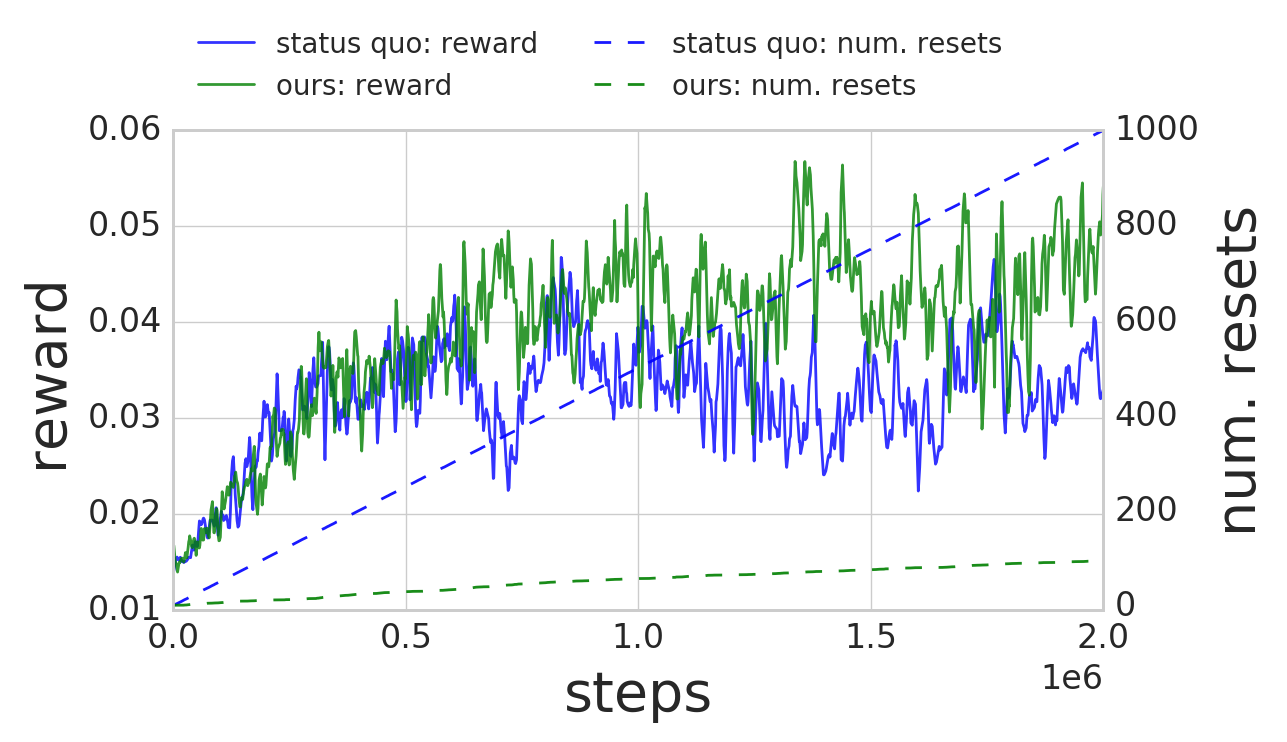}
        \caption*{Cliff Cheetah}
    \end{subfigure}%
    \caption{Our method achieves equal or better rewards than the status quo with fewer manual resets. \label{fig:learning-with-resets}}
\end{figure}

Our first goal is to reduce the number of hard resets during learning. In this section, we compare our algorithm to the standard, episodic learning setup (``status quo''), which only learns a forward policy.
As shown in Figure~\ref{fig:learning-with-resets} (left), the conventional approach learns the pusher task somewhat faster than ours, but our approach eventually achieves the same reward with half the number of hard resets. In the cliff cheetah task (Figure~\ref{sec:learning-with-resets} (right)), not only does our approach use an order of magnitude fewer hard resets, but the final reward of our method is substantially higher. This suggests that, besides reducing the number of resets, the early aborts can actually aid learning by preventing the forward policy from wasting exploration time waiting for resets in irreversible states.

\subsection{Do Early Aborts avoid Hard Resets?}
\label{sec:early-aborts-experiment}

\begin{figure}[th!]
    \centering
    \begin{subfigure}[t]{0.5\textwidth}
        \centering
        \includegraphics[width=\textwidth]{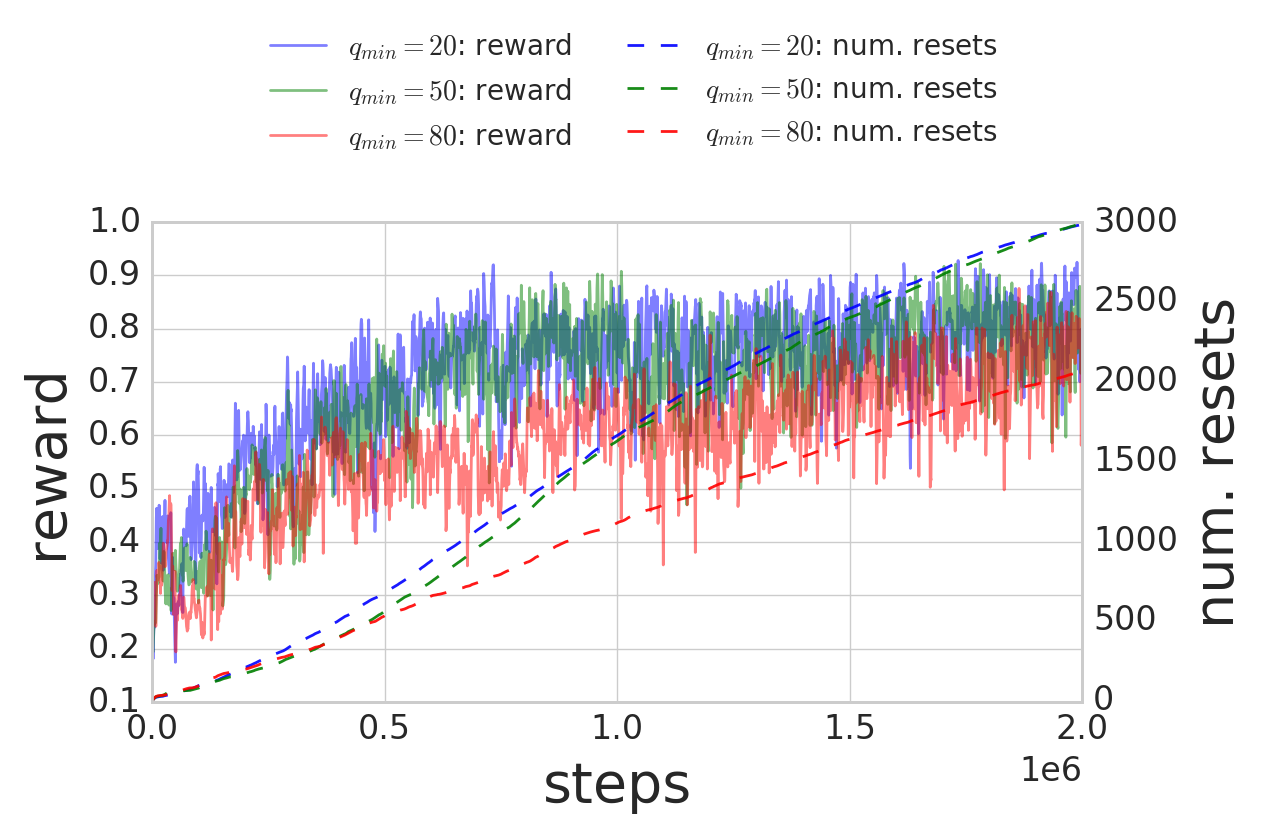}
        \caption*{Pusher}
    \end{subfigure}%
     ~ 
    \begin{subfigure}[t]{0.5\textwidth}
        \centering
        \includegraphics[width=\textwidth]{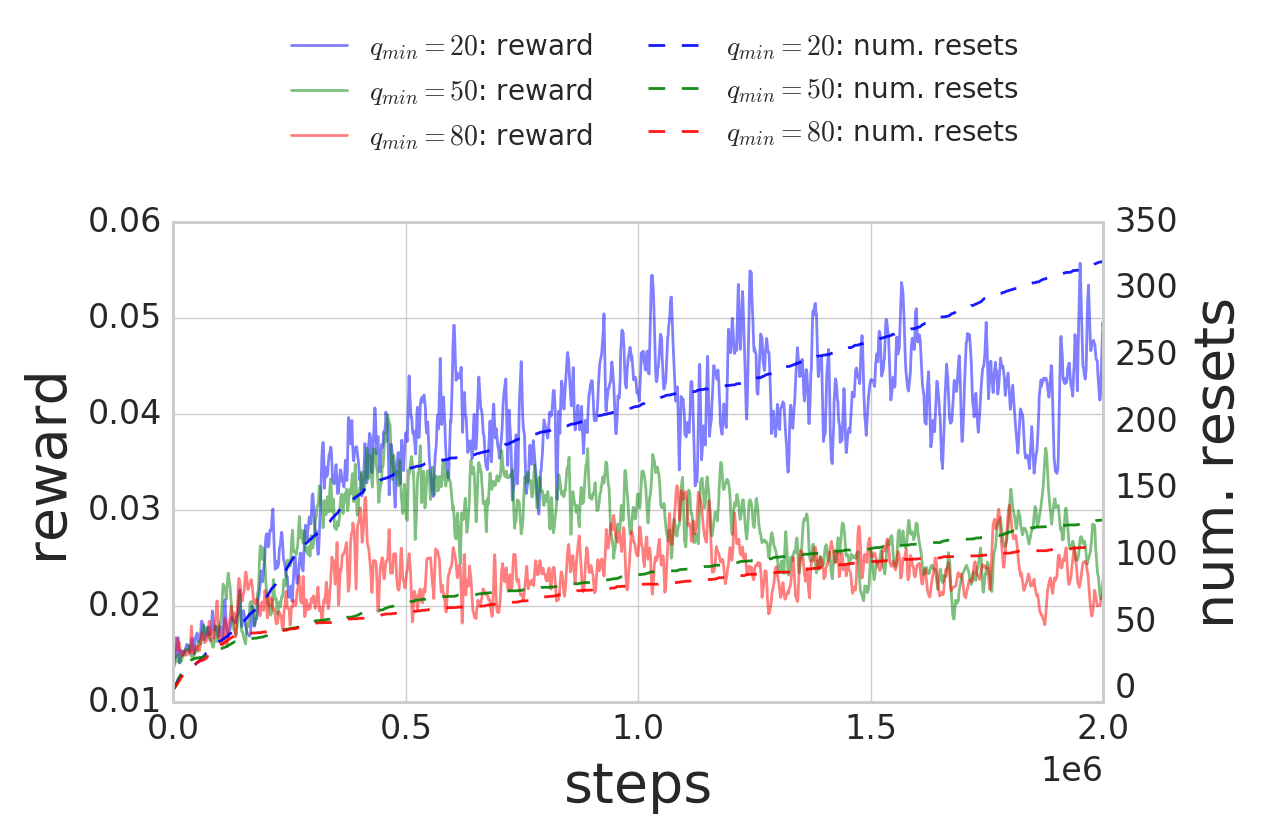}
        \caption*{Cliff Cheetah}
    \end{subfigure}%
    \caption{Early abort threshold: Increasing the early abort threshold to act more cautiously avoids many hard resets, indicating that early aborts help avoid irreversible states. \label{fig:early-aborts-experiment}}
\end{figure}

To test whether early aborts prevent hard resets, we can see if the number of hard resets increases when we lower the early abort threshold. Figure~\ref{fig:early-aborts-experiment} shows the effect of three values for $Q_{min}$ while learning the pusher and cliff cheetah. In both environments, decreasing the early abort threshold increased the number of hard resets, supporting our hypothesis that early aborts prevent hard resets. On pusher, increasing $Q_{min}$ to 80 allowed the agent to learn a policy that achieved nearly the same reward using 33\% fewer hard resets. The cliff cheetah task has lower rewards than pusher, even an early abort threshold of 10 is enough to prevent 69\% of the total early aborts that the status quo would have performed.

\subsection{Multiple Reset Attempts}
\label{sec:reset-attempts-experiment}

While early aborts help avoid hard resets, our algorithm includes a mechanism for requesting a manual reset if the agent reaches an unresettable state. As described in Section~\ref{sec:manual-resets}, we only perform a hard reset if the reset agent fails to reset in $N$ consecutive episodes.

\begin{figure}[H]
    \centering
    \begin{subfigure}[t]{0.5\textwidth}
        \centering
        \includegraphics[width=\textwidth]{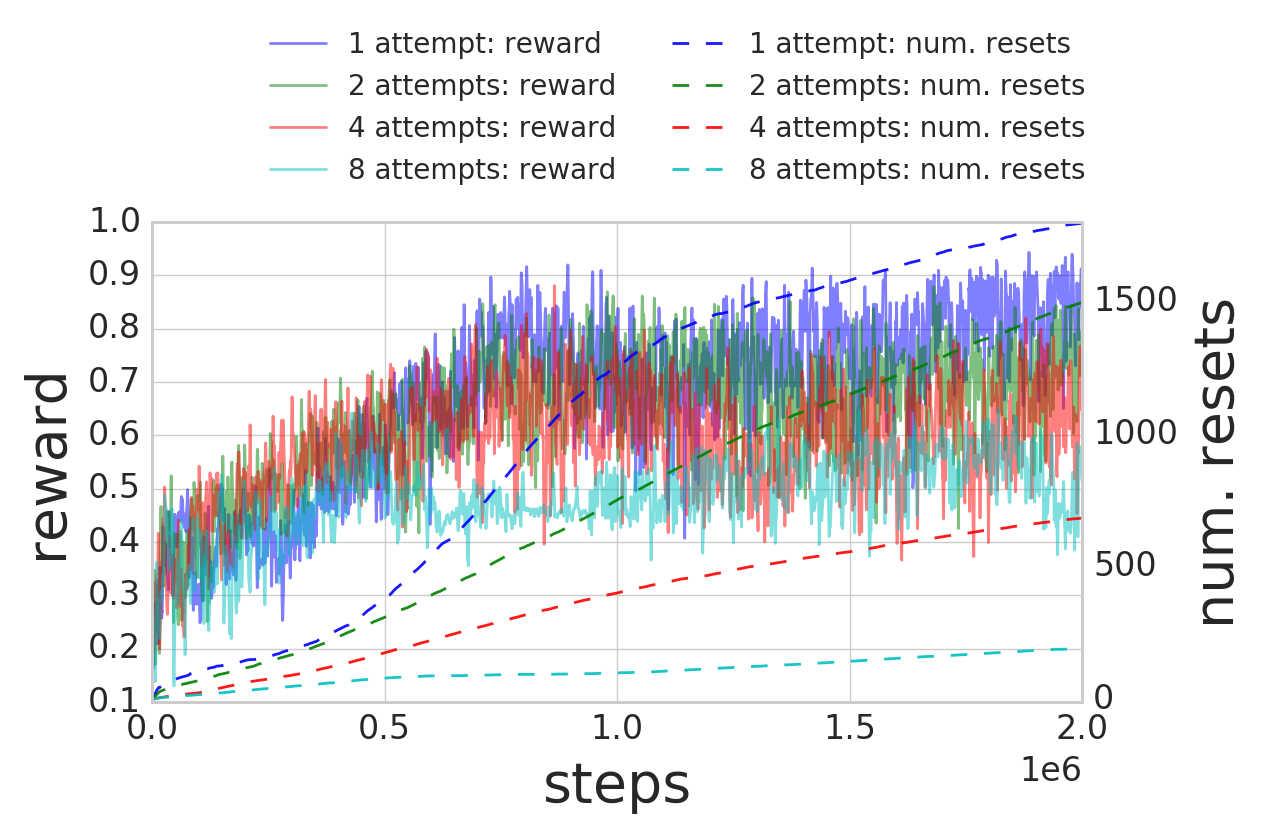}
        \caption*{Pusher}
    \end{subfigure}%
     ~ 
    \begin{subfigure}[t]{0.5\textwidth}
        \centering
        \includegraphics[width=\textwidth]{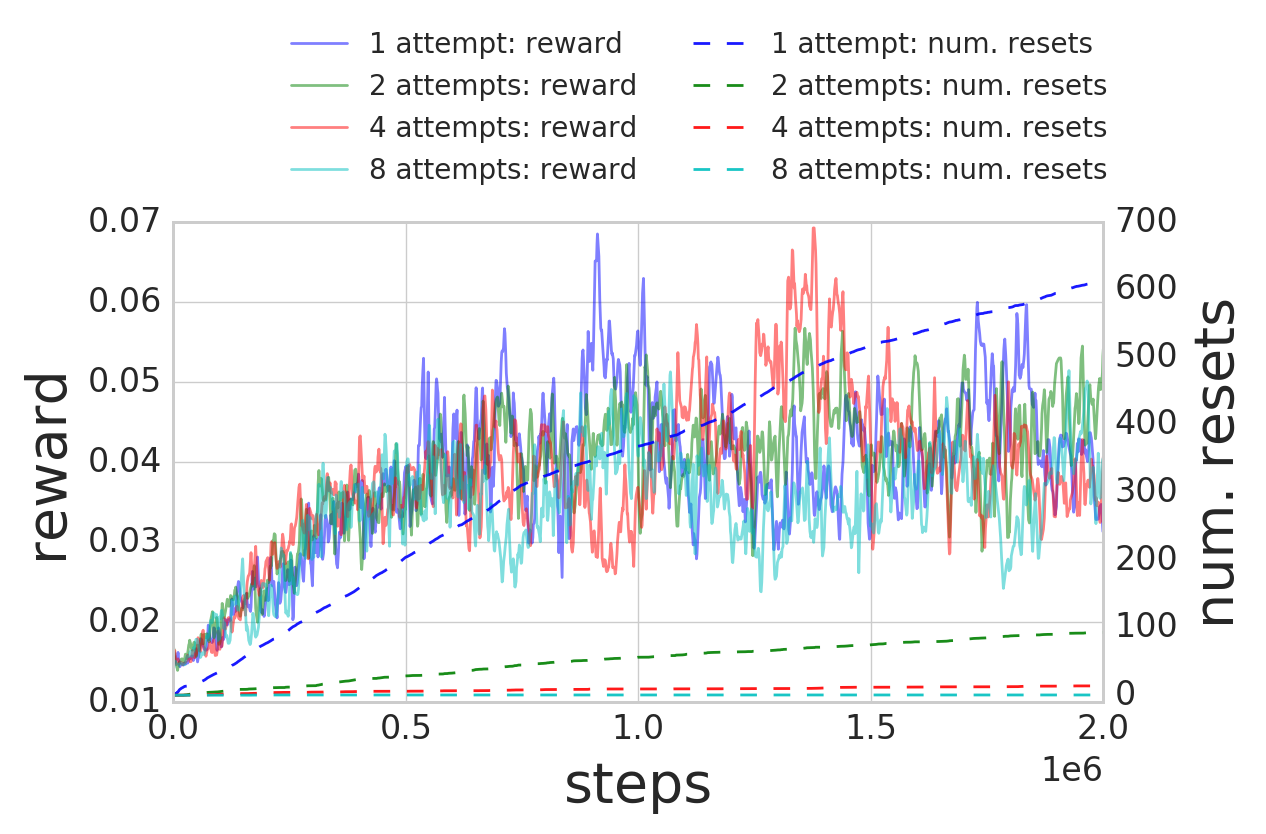}
        \caption*{Cliff Cheetah}
    \end{subfigure}%
    \caption{Reset attempts: Increasing the number of reset attempts reduces hard resets. Allowing too many reset attempts reduces reward for the pusher environment. \label{fig:reset-attempts-experiment}}
\end{figure}

Figure~\ref{fig:reset-attempts-experiment} shows how the number of reset attempts, $N$, affects hard resets and reward.
On the pusher task, when our algorithm was given a single reset attempt, it used 64\% fewer hard resets than the status quo approach would have. Increasing the number of reset attempts to 4 resulted in another 2.5x reduction in hard resets, while decreasing the reward by less than 25\%.
On the cliff cheetah task, increasing the number of reset attempts brought the number of resets down to nearly 0, without changing the reward. Surprisingly, these results indicate that for some tasks, it is possible to learn an equally good policy with significantly fewer hard resets.

\subsection{Ensembles are Safer}
\label{sec:ensemble-experiment}

\begin{wrapfigure}[14]{r}{0.45\textwidth}
    \captionsetup{width=0.4\textwidth}
    \vspace{-2.5em}
    \centering
        \includegraphics[width=0.5\textwidth]{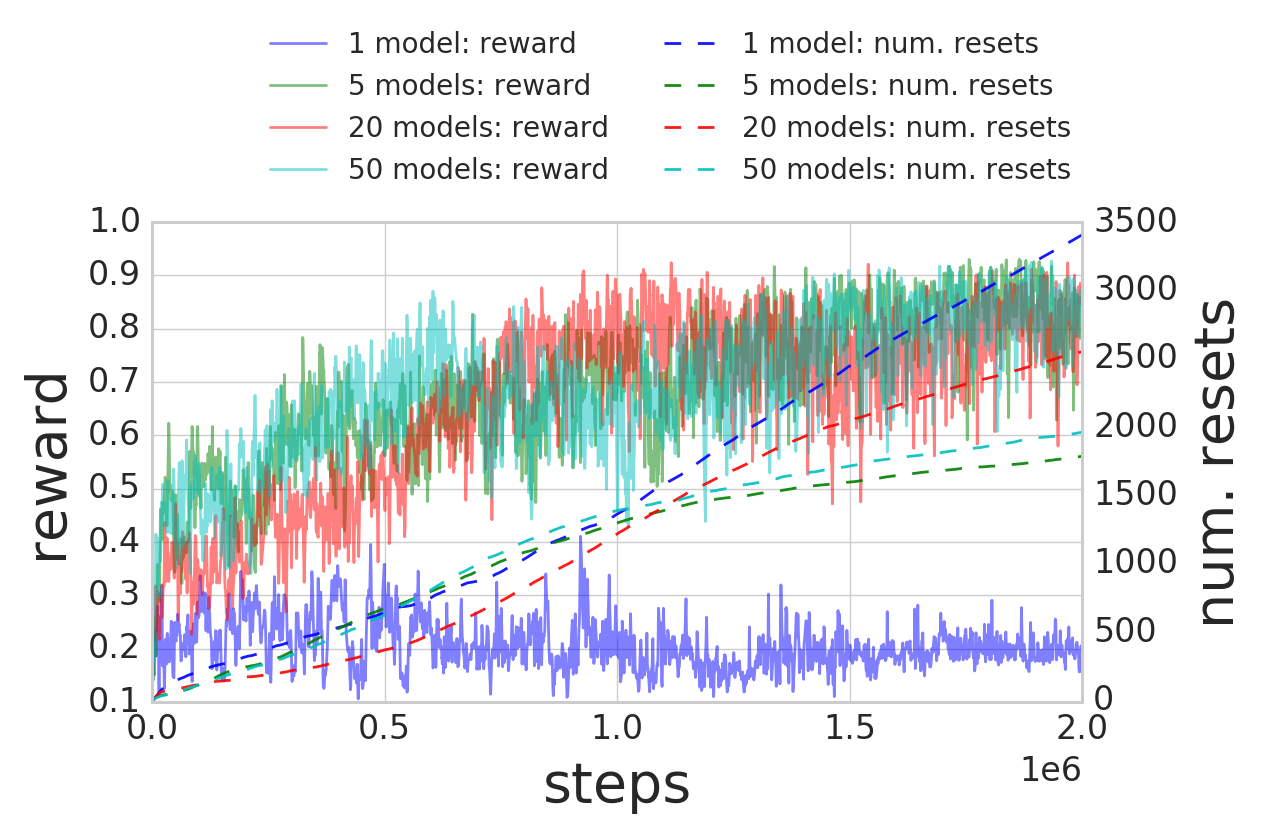}
        \vspace{-1.5em}
        \caption{Increasing ensemble size boosts policy reward while decreasing rate of hard resets. \label{fig:ensemble-experiment}}
\end{wrapfigure}

Our approach uses an ensemble of value functions to trigger early aborts. Our hypothesis was that our algorithm would be sensitive to bias in the value function if we used a single Q network. To test this hypothesis, we varied the ensemble size from 1 to 50. Figure~\ref{fig:ensemble-experiment} shows the effect on learning the pushing task. An ensemble with one network failed to learn, but still required many hard resets. Increasing the ensemble size slightly decreased the number of hard resets without affecting the reward.

\subsection{Automatic Curriculum Learning}
\label{sec:curriculum-learning}

Our method can automatically produce a curriculum in settings where the desired skill is performed by the reset policy, rather than the forward policy. As an example, we evaluate our method on a peg insertion
\vspace{-1\parskip}

\begin{wrapfigure}[12]{r}{0.45\textwidth}
    \captionsetup{width=0.4\textwidth}
    \vspace{-4.5em}
    \centering
        \includegraphics[width=0.5\textwidth]{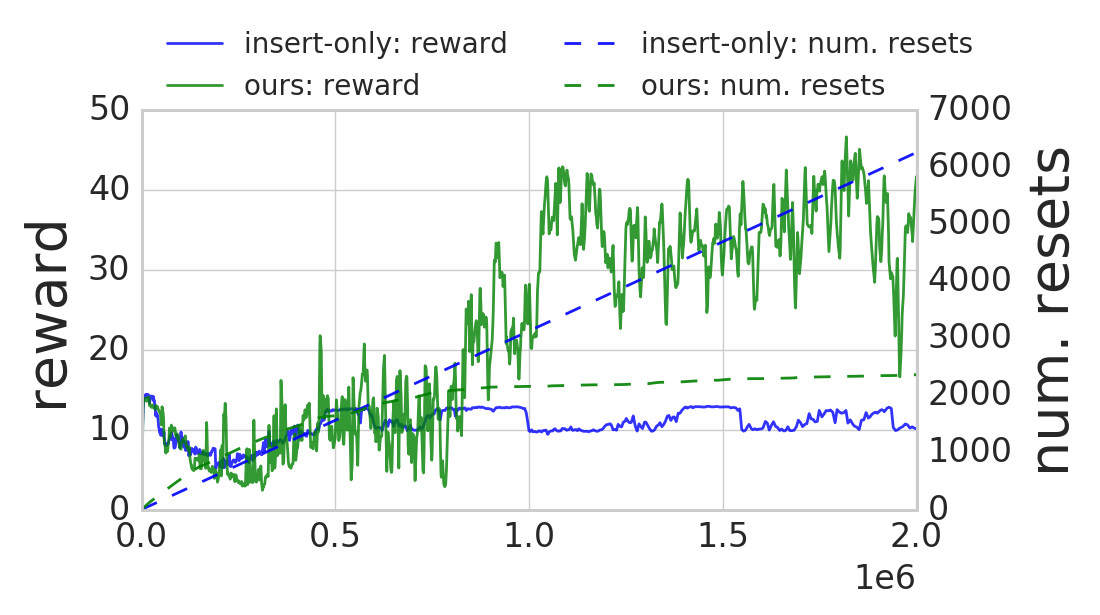}
        \vspace{-1.5em}
        \caption{Our method automatically induces a curriculum, allowing the agent to solve peg insertion with sparse rewards. \label{fig:curriculum-learning}}
\end{wrapfigure}

task, where the reset policy inserts the peg and the forward policy removes it.
The reward for a successful peg insertion is provided only when the peg is in the hole, making this task challenging to learn with random exploration. Hard resets provide illustrations of what a successful outcome looks like, but do not show how to achieve it.
Our algorithm starts with the peg in the hole and runs the forward (peg removal) policy until an early abort occurs. As the reset (peg insertion) policy improves, early aborts occur further and further from the hole. Thus, the initial state distribution for the reset (peg insertion) policy moves further and further from the hole, increasing the difficulty of the task as the policy improves.
We compare our approach to an ``insert-only'' baseline that only learns the peg insertion policy -- we manually remove the peg from the hole after every episode. For evaluation, both approaches start outside the hole.
Figure~\ref{fig:curriculum-learning} shows that only our method solves the task. The number of resets required by our method plateaus after one million steps, indicating that it has solved the task and no longer requires hard resets at the end of the episode. In contrast, the ``insert-only'' baseline fails to solve the task, never improving its reward. Thus, even if reducing manual resets is not important, the curriculum automatically created by Leave No Trace can enable agents to learn policies they otherwise would be unable to solve.

\section{Conclusion}
\label{sec:conclusion}

In this paper, we presented a framework for automating reinforcement learning based on two principles: automated resets between trials, and early aborts to avoid unrecoverable states. Our method simultaneously learns a forward and reset policy, with the value functions of the two policies used to balance exploration against recoverability.
Experiments in this paper demonstrate that our algorithm not only reduces the number of manual resets required to learn a task, but also learns to avoid unsafe states and automatically induces a curriculum.

Our algorithm can be applied to a wide range of tasks, only requiring a small number of manual resets to learn some tasks.
During the early stages of learning we cannot accurately predict the consequences of our actions. We cannot learn to avoid a dangerous state until we have visited that state (or a similar state) and experienced a manual reset. Nonetheless, reducing the number of manual resets during learning will enable researchers to run experiments for longer on more agents.
A second limitation of our work is that we treat all manual resets as equally bad. In practice, some manual resets are more costly than others. For example, it is more costly for a grasping robot to break a wine glass than to push a block out of its workspace. An approach not studied in this paper for handling these cases would be to specify costs associated with each type of manual reset, and incorporate these reset costs into the learning algorithm.

While the experiments for this paper were done in simulation, where manual resets are inexpensive, the next step is to apply our algorithm to real robots, where manual resets are costly. A challenge introduced when switching to the real world is automatically identifying when the agent has reset. In simulation we can access the state of the environment directly to compute the distance between the current state and initial state. In the real world, we must infer states from noisy sensor observations to deduce if they are the same.  If we cannot distinguish between the state where the forward policy started and the state where the reset policy ended, then we have succeeded in Leaving No Trace!

\vspace{4em}
\textbf{Acknowledgements:} We thank Sergio Guadarrama, Oscar Ramirez, and Anoop Korattikara for implementing DDPG and thank Peter Pastor for insightful discussions.

{\small

\bibliography{iclr2018_conference}
\bibliographystyle{iclr2018_conference}
}

\clearpage
\appendix

\section{Combining an Ensemble of Value Functions}
\label{sec:combining-ensemble-experiments}

\begin{figure}[th!]
    \centering
    \vspace{-1em}
    \begin{subfigure}[t]{0.4\textwidth}
        \centering
        \includegraphics[width=\textwidth]{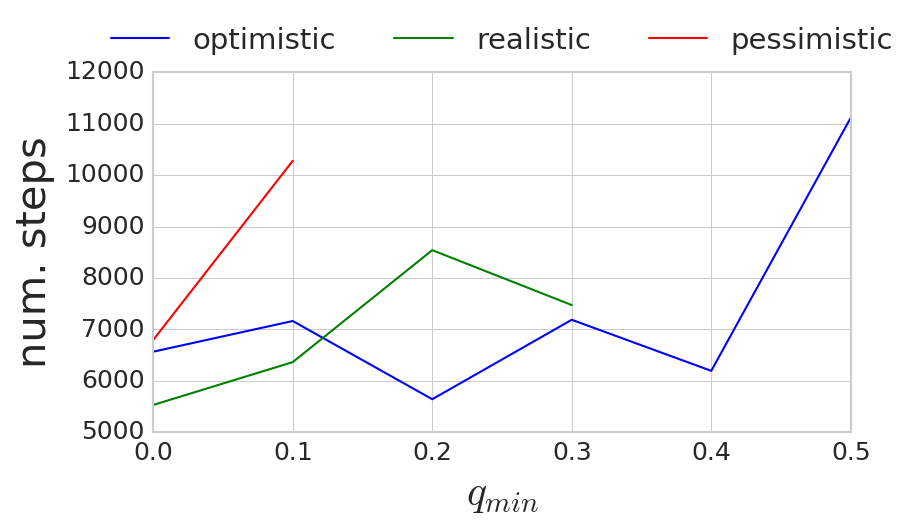}
    \end{subfigure}%
    ~
    \begin{subfigure}[t]{0.4\textwidth}
        \centering
        \includegraphics[width=\textwidth]{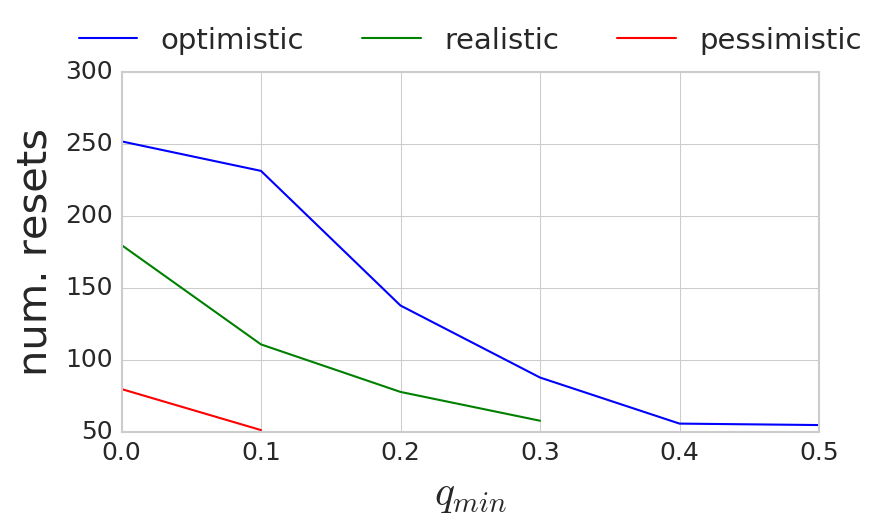}
    \end{subfigure}
    \vspace{-0.5em}
    \caption{Combining value functions: We compare three methods for ensembling value functions on gridworld. Missing points for the red and green lines indicate that pessimistic and realistic method fail to solve the task for larger values of $Q_{min}$. \label{fig:ensemble-method-gridworld}}
    \vspace{-0.5em}
\end{figure}

We benchmarked three methods for combining our ensemble of values functions (optimistic, realistic, and pessimistic, as discussed in Section~\ref{sec:ensembles}). Figure~\ref{fig:ensemble-method-gridworld} compares the three methods on gridworld on the gridworld environment from Section~\ref{sec:gridworld-experiments}. Only the optimistic agent efficiently explored.  As expected, the realistic and pessimistic agents, which are more conservative in letting the forward policy continue, fail to explore when $Q_{min}$ is too large.

\begin{figure}[th!]
    \centering
    \vspace{-0.5em}
    \begin{subfigure}[t]{0.33\textwidth}
        \centering
        \includegraphics[width=\textwidth]{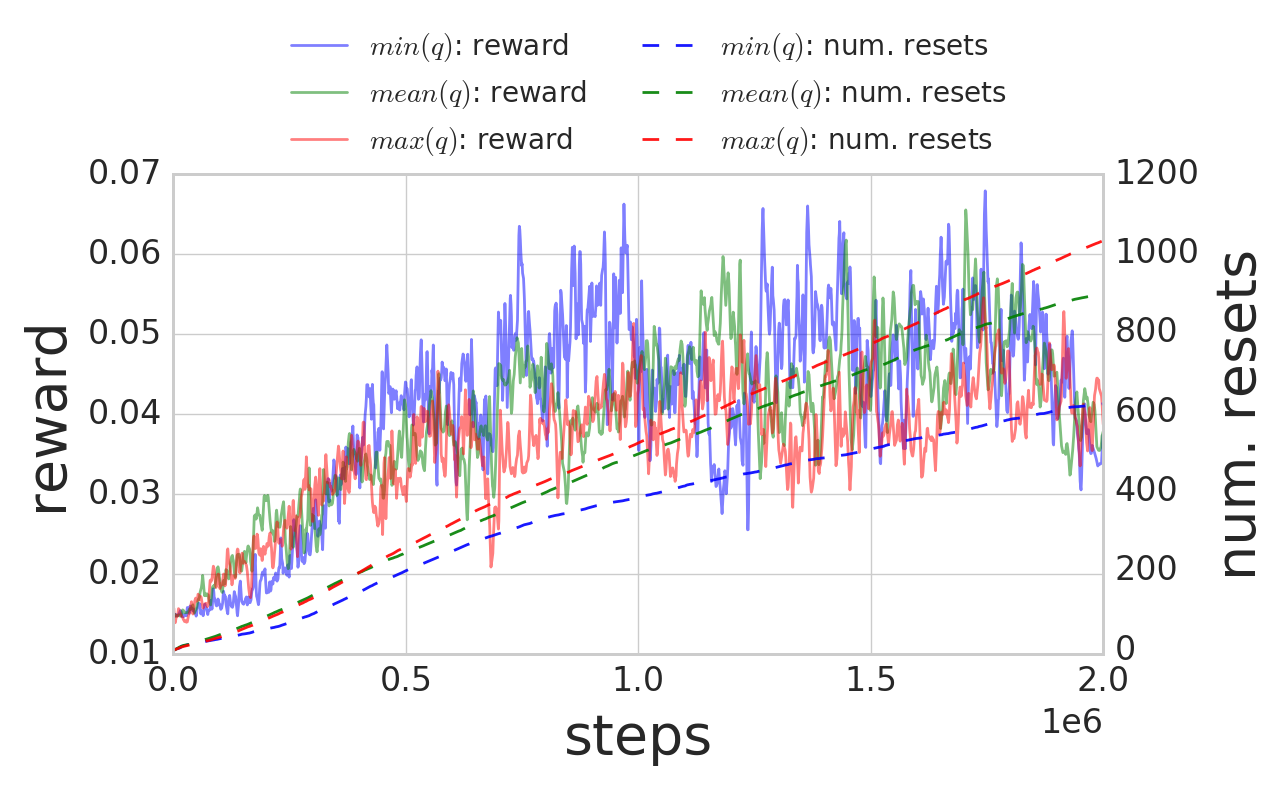}
        \caption{Cliff Cheetah}
    \end{subfigure}%
    ~
    \begin{subfigure}[t]{0.33\textwidth}
        \centering
        \includegraphics[width=\textwidth]{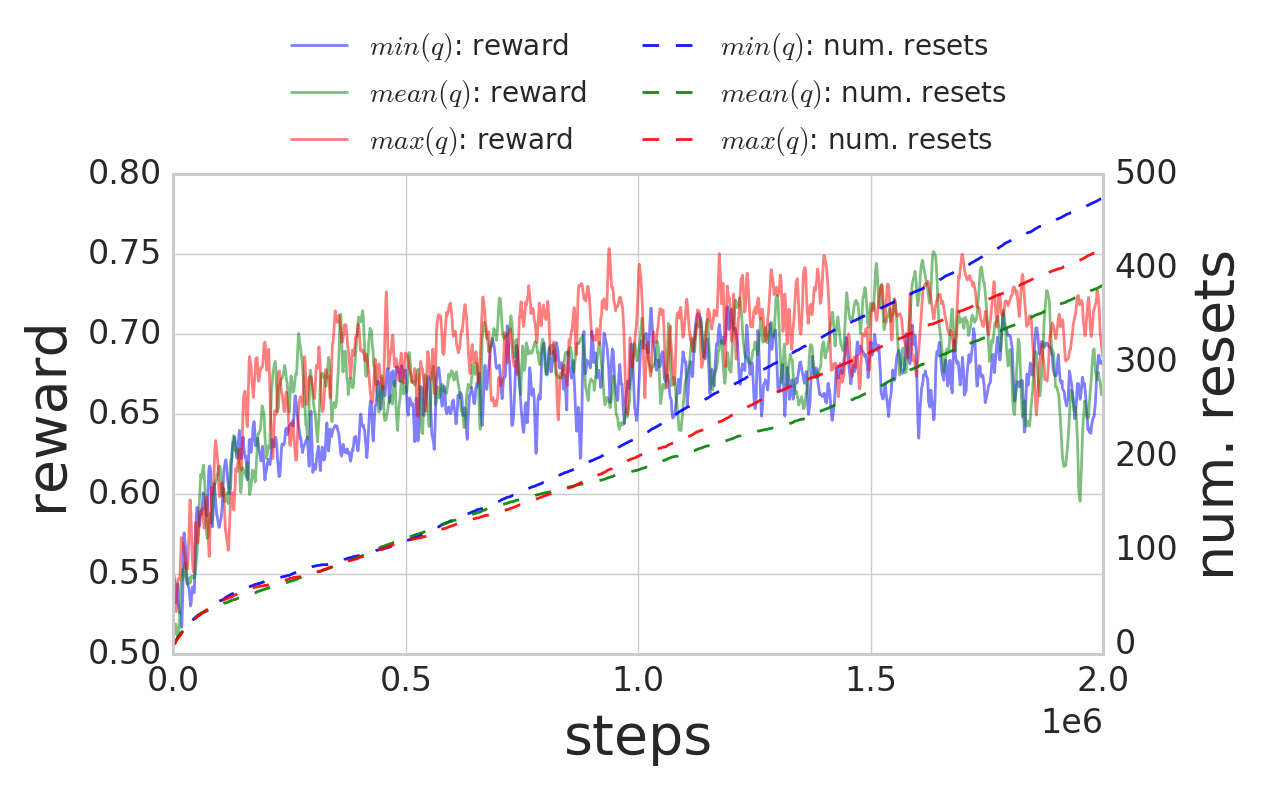}
        \caption{Cliff Walker}
    \end{subfigure}%
    ~
    \begin{subfigure}[t]{0.33\textwidth}
        \centering
        \includegraphics[width=\textwidth]{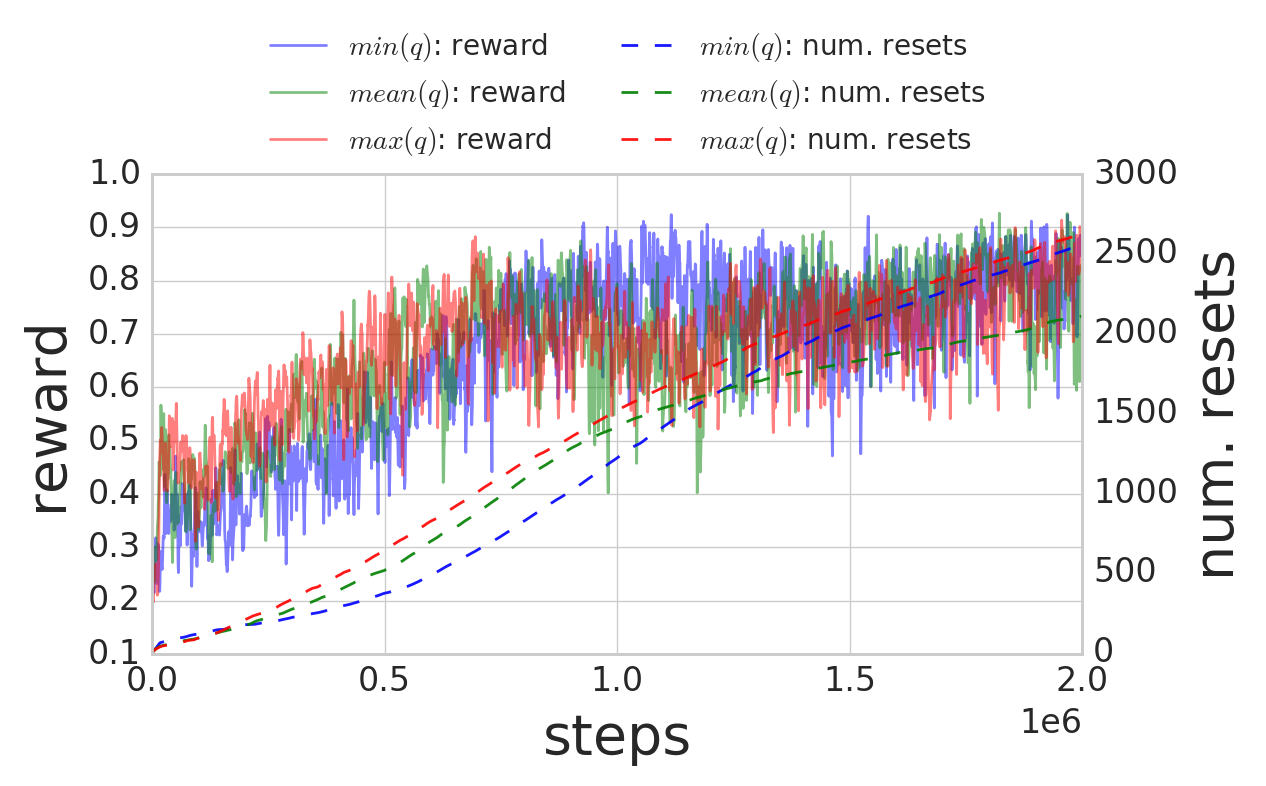}
        \caption{Pusher}
    \end{subfigure}%
    \caption{Combining value functions: For continuous environments, the method for combing value functions has little effect. \label{fig:ensemble-method-cts}}
    \vspace{-1em}
\end{figure}

Interestingly, for the continuous control environments, the ensembling method makes relatively little difference for the number of resets or final performance, as shown in Figure~\ref{fig:ensemble-method-cts}. This suggests that much of the benefit of ensemble comes from its ability to produce less biased abort predictions in novel states, rather than the particular risk-sensitive rule that is used. Additionally, the agent's value function may generalize better over continuous state spaces compared with the tabular gridworld.

\section{Additional Figures}
\label{appendix:plots}
For each experiment in the main paper, we chose one or two demonstrative environments. Below, we show all experiments run on cliff cheetah, cliff walker, and pusher.

\subsection{Does Our Method Reduce Manual Resets? -- More Plots}
\begin{figure}[H]
    \centering
    \begin{subfigure}[t]{0.33\textwidth}
        \centering
        \includegraphics[width=\textwidth]{ours_vs_convention_n_4_cheetah_cliff.png}
        \caption{Cliff Cheetah}
    \end{subfigure}%
    ~
    \begin{subfigure}[t]{0.33\textwidth}
        \centering
        \includegraphics[width=\textwidth]{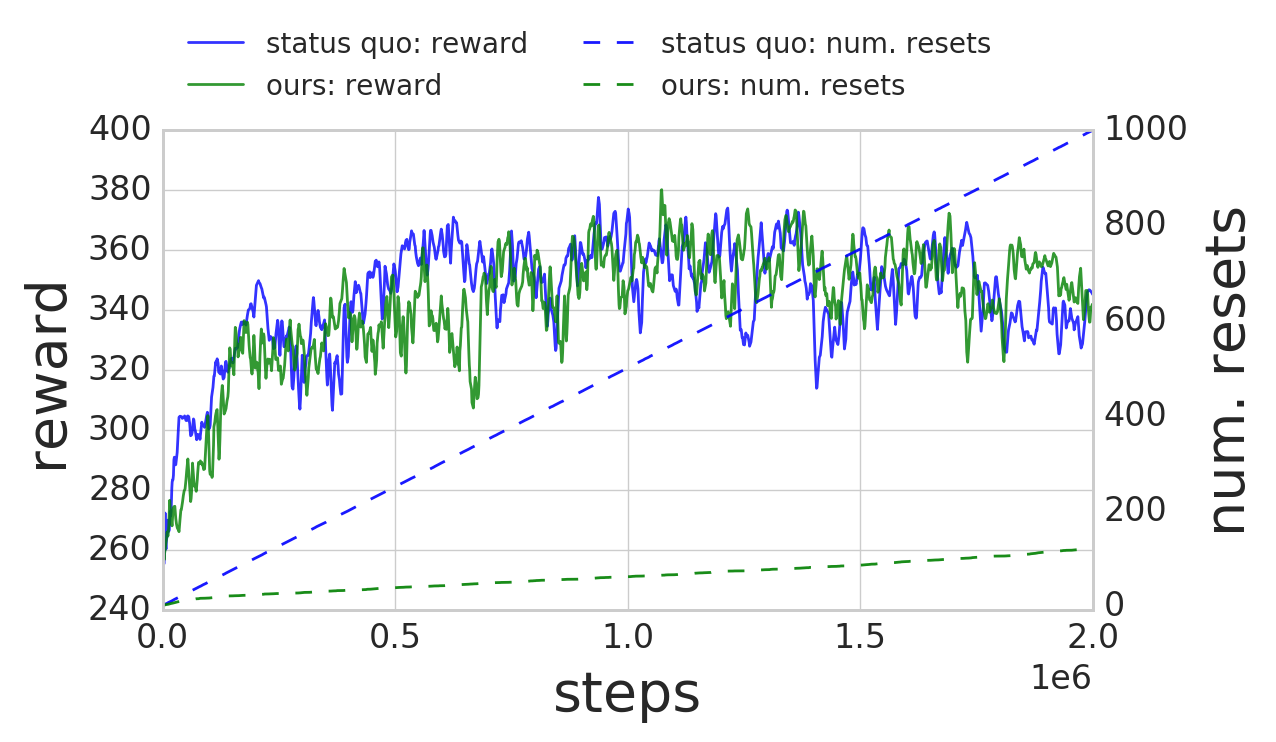}
        \caption{Cliff Walker}
    \end{subfigure}%
    ~
    \begin{subfigure}[t]{0.33\textwidth}
        \centering
        \includegraphics[width=\textwidth]{ours_vs_convention_n_4_pusher_multi_reward.png}
        \caption{Pusher}
    \end{subfigure}%
    \caption{Experiment from \S~\ref{sec:learning-with-resets} \label{fig:learning-with-resets-extra}}
\end{figure}

This experiment, described in Section~\ref{sec:learning-with-resets}, compared our method to the status quo approach (resetting after every episode). Figure~\ref{fig:learning-with-resets-extra} shows plots for all environments.

\subsection{Do Early Aborts avoid Hard Resets Plots? -- More Plots}
\begin{figure}[H]
    \centering
    \begin{subfigure}[t]{0.33\textwidth}
        \centering
        \includegraphics[width=\textwidth]{early_aborts_cheetah_cliff.png}
        \caption{Cliff Cheetah}
    \end{subfigure}%
    ~
    \begin{subfigure}[t]{0.33\textwidth}
        \centering
        \includegraphics[width=\textwidth]{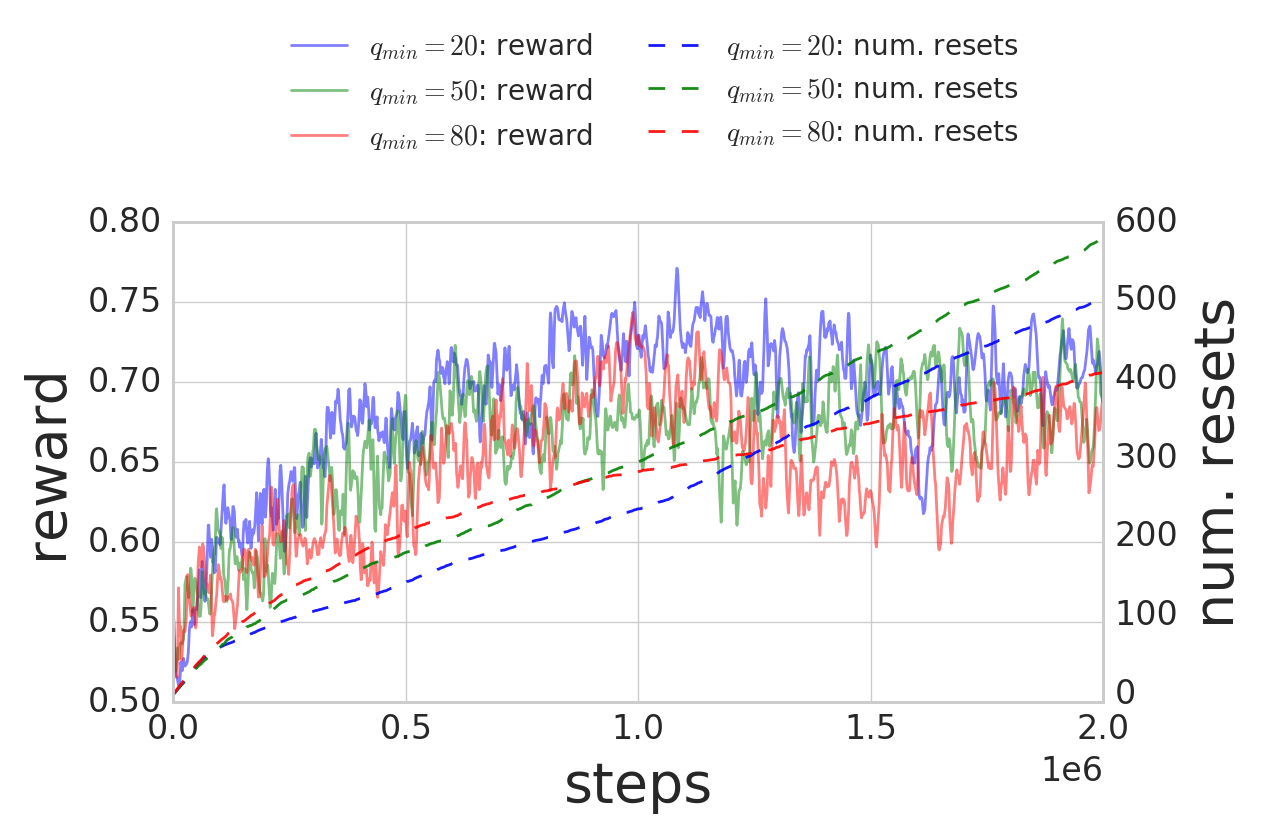}
        \caption{Cliff Walker}
    \end{subfigure}%
    ~
    \begin{subfigure}[t]{0.33\textwidth}
        \centering
        \includegraphics[width=\textwidth]{early_aborts_pusher.png}
        \caption{Pusher}
    \end{subfigure}%
    \caption{Experiment from \S~\ref{sec:early-aborts-experiment} \label{fig:early-aborts-experiment-extra}}
\end{figure}

This experiment, described in Section~\ref{sec:early-aborts-experiment}, shows the effect of varying the early abort threshold. Figure~\ref{fig:early-aborts-experiment-extra} shows plots for all environments.

\subsection{Multiple Reset Attempts -- More Plots}
\begin{figure}[H]
    \centering
    \begin{subfigure}[t]{0.33\textwidth}
        \centering
        \includegraphics[width=\textwidth]{reset_attempts_cheetah_cliff.png}
        \caption{Cliff Cheetah}
    \end{subfigure}%
    ~
    \begin{subfigure}[t]{0.33\textwidth}
        \centering
        \includegraphics[width=\textwidth]{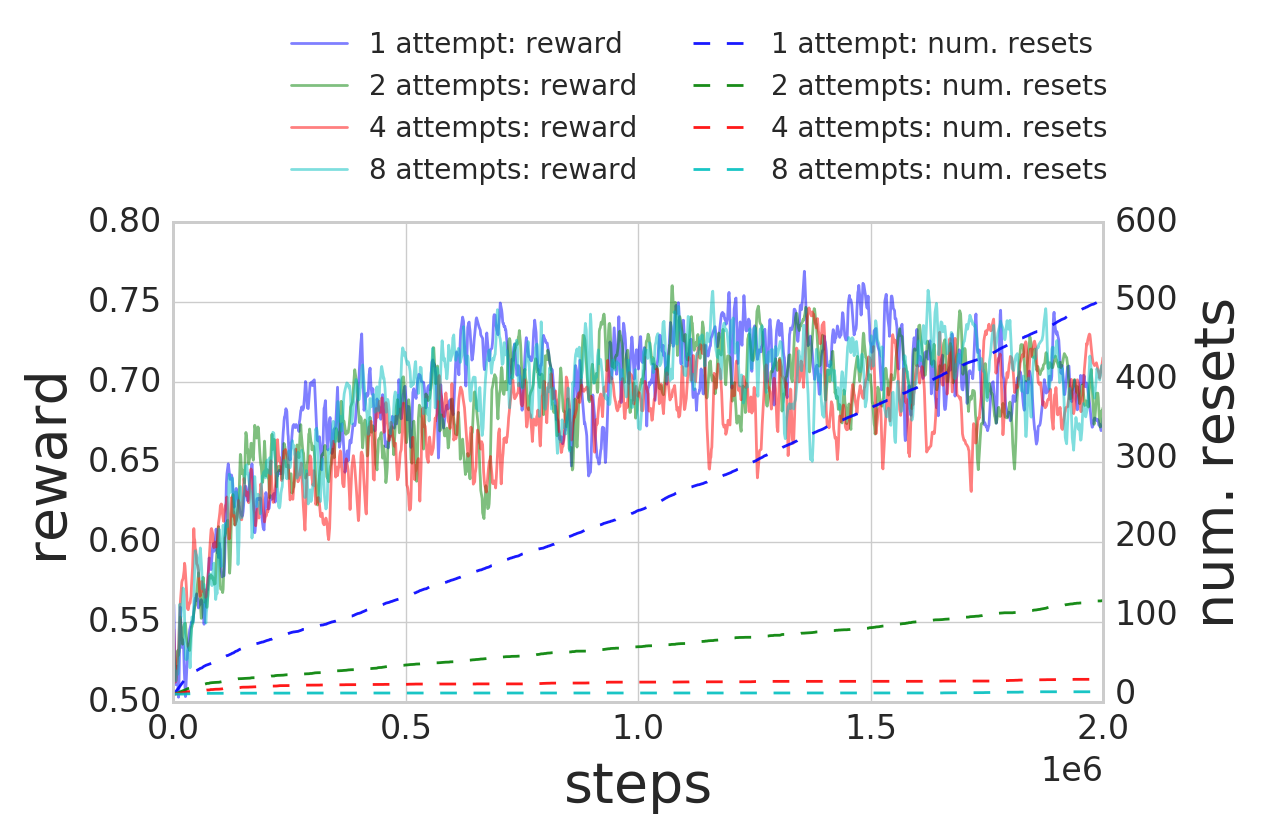}
        \caption{Cliff Walker}
    \end{subfigure}%
    ~
    \begin{subfigure}[t]{0.33\textwidth}
        \centering
        \includegraphics[width=\textwidth]{reset_attempts_pusher_multi_reward.png}
        \caption{Pusher}
    \end{subfigure}%
    \caption{Experiment from \S~\ref{sec:reset-attempts-experiment} \label{fig:reset-attempts-experiment-extra}}
\end{figure}

This experiment, described in Section~\ref{sec:reset-attempts-experiment}, shows the effect of increasing the number of reset attempts. Figure~\ref{fig:reset-attempts-experiment-extra} shows plots for all environments.

\subsection{Ensembles are Safer -- More Plots}
\begin{figure}[H]
    \centering
    \begin{subfigure}[t]{0.33\textwidth}
        \centering
        \includegraphics[width=\textwidth]{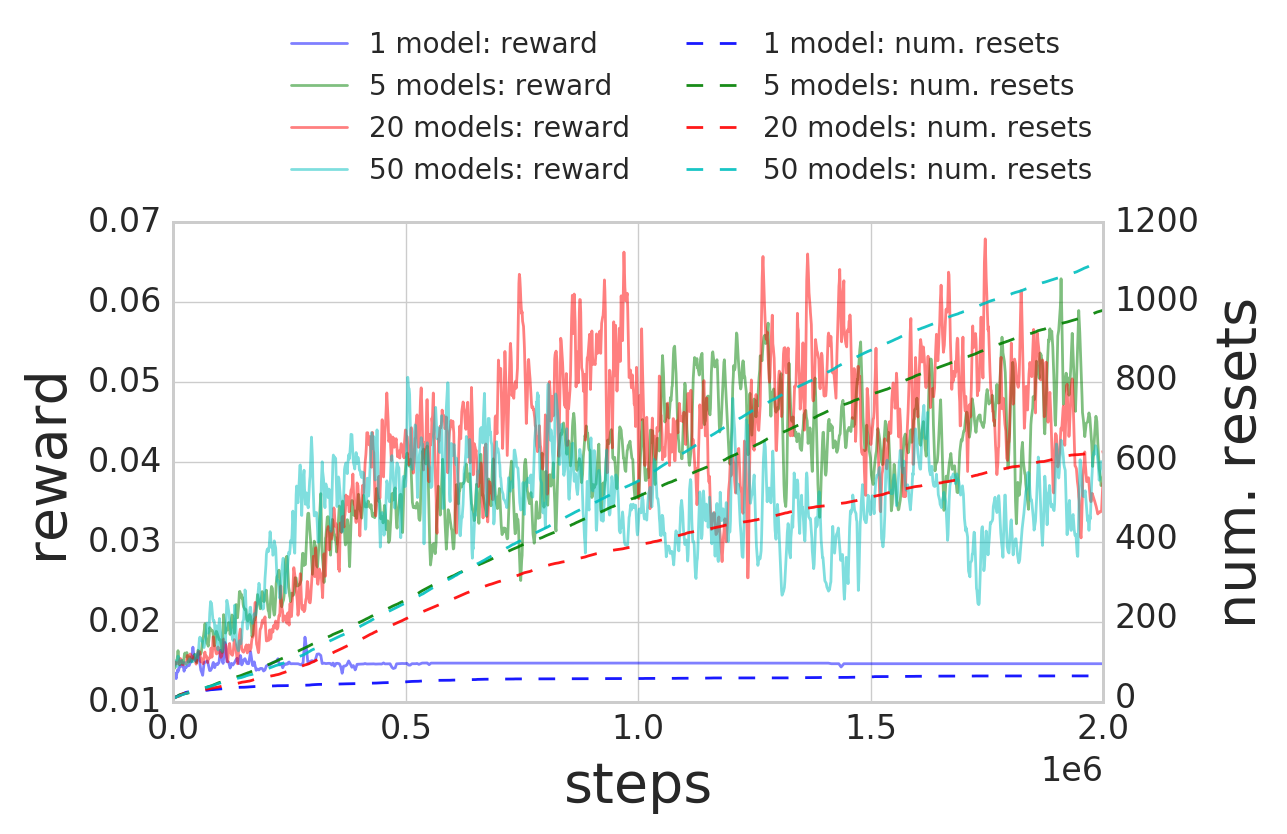}
        \caption{Cliff Cheetah}
    \end{subfigure}%
    ~
    \begin{subfigure}[t]{0.33\textwidth}
        \centering
        \includegraphics[width=\textwidth]{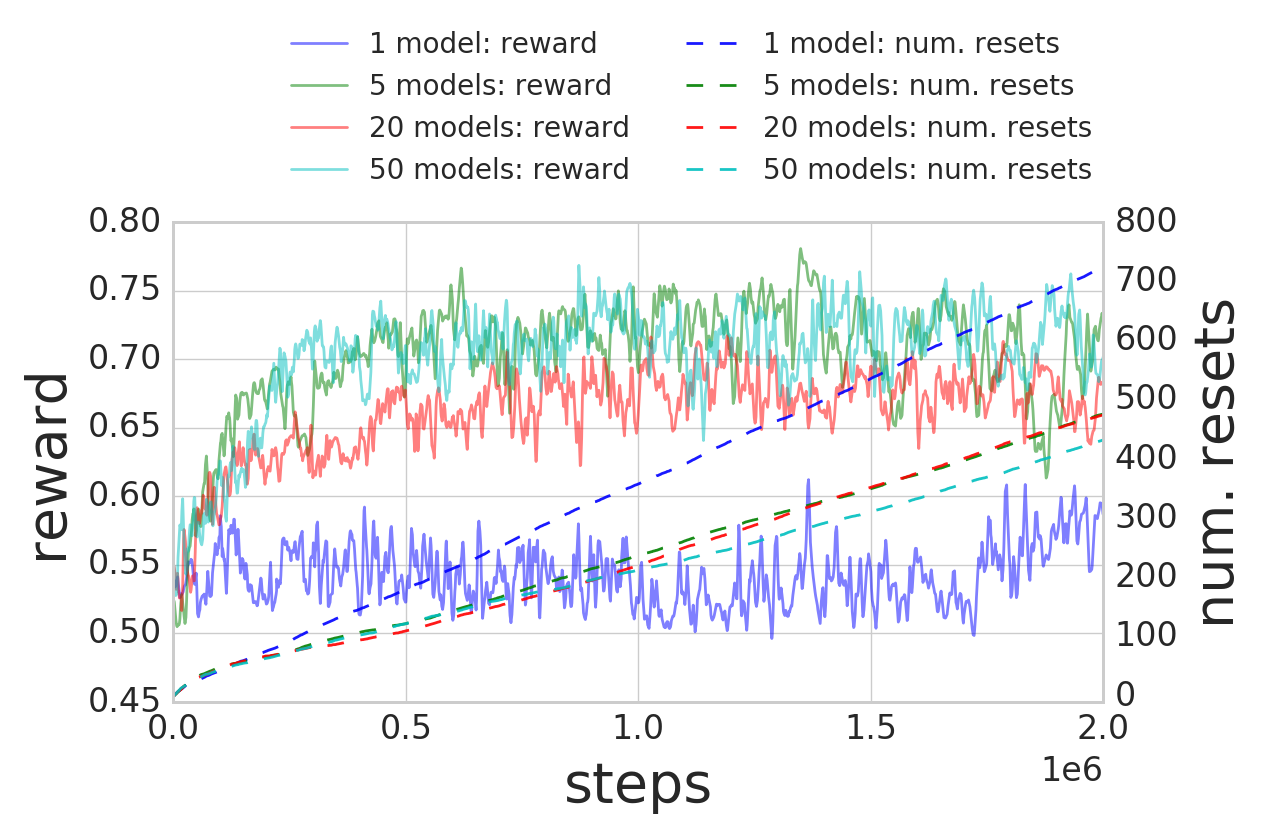}
        \caption{Cliff Walker}
    \end{subfigure}%
    ~
    \begin{subfigure}[t]{0.33\textwidth}
        \centering
        \includegraphics[width=\textwidth]{ensemble_size_pusher_multi_reward.png}
        \caption{Pusher}
    \end{subfigure}%
    \caption{Experiment from \S~\ref{sec:ensemble-experiment} \label{fig:ensemble-experiment-extra}}
\end{figure}

This experiment, described in Section~\ref{sec:ensemble-experiment}, shows the effect of increasing the number of reset attempts. Figure~\ref{fig:ensemble-experiment-extra} shows plots for all environments.

\clearpage

\section{Experimental Details}
\label{appendix:experiments}

\subsection{Gridworld Experiments}
To generate Figures~\ref{fig:where-are-early-aborts} and \ref{fig:where-are-early-aborts-irreversible}, we averaged early abort counts across across 10 random seeds. For Figure~\ref{fig:gridworld-qmin} we took the median result across 10 random seeds. Both gridworld experiments used 5 models in the ensemble.

\subsection{Continuous Control Environments}
\begin{itemize}
    \item[] \emph{Ball in Cup:} The agent receives a reward of 1 if the ball is in the cup and 0 otherwise. We defined the reward for the reset task to be the negative distance between current state and the initial state (ball hanging stationary below the cup).
    \item[] \emph{Cliff Cheetah:} The agent learns how to run on a 14m cliff. The agent is rewarded for moving forward. We defined the reset reward to be a combination of the distance from the origin, an indicator of whether the agent was standing, and a control penalty.
    \item[] \emph{Cliff Walker:} The agent learns how to walker on a 6m cliff. The agent is rewarded for moving forward. We defined the reset reward to be a combination of the distance from the origin, an indicator of whether the agent was standing, and a control penalty.
    \item[] \emph{Pusher:} The agent pushes a puck to a goal location. The agent's reward is a combination of the distance from the puck to the goal, the distance from arm to the puck, and a control penalty. For the reset reward, we use the distance from puck to start instead of distance from puck to goal.
    \item[] \emph{Peg Insertion:} The agent inserts a peg into a small hole. For the insertion task, the reward is 1 if the peg is in the hole and 0 otherwise. We add a control penalty to stabilize learning. For the peg removal task, the reward is the negative distance of the state to a fixed state outside the hole, in addition to a control penalty.
\end{itemize}

\subsection{Continuous Control Experiments}

We did not do hyperparameter optimization for our experiments, but did run with 5 random seeds. To aggregate results, we took the median number across all random seeds that solved the task. For most experiments, all random seeds solved the task.

For the three continuous control environments, we normalized the rewards to be in $[0, 1]$ so we could use the same hyperparameters for each. Using a discount factor of $\gamma = 0.99$, the cumulative discounted reward was in $[0, 100)$. We defined $\mathcal{S}_{reset}$ as states with reset reward was greater than 0.7.

We used the same DDPG hyperparameters for all continuous control environments:
\begin{itemize}
    \item[] \emph{Actor Network:} Two fully connected layers of sizes 400 and 300, with tanh nonlinearities throughout.
    \item[] \emph{Critic Network:} We apply a 400-dimensional fully connected layer to states, then concatenate the actions and apply another 300-dimensional fully connected layer. Again, we use tanh nonlinearities.
\end{itemize}

Unless otherwise noted, experiments used an ensemble of size 20, $Q_{min} = 10$, 1 reset attempt, and early aborts using $min(q)$. The experiments in Section~\ref{sec:learning-with-resets}, our model used 2 reset attempts to better illustrate the potential for our approach to reduce hard resets.

\end{document}